# Prion-ViT: Prions-Inspired Vision Transformers for Temperature Prediction with Specklegrams

Abhishek Sebastian[1,2][0000-0002-3241-1450], Pragna R[1][0000-0003-0827-5896], Sonaa Rajagopal[1][0009-0001-4572-6863], and Muralikrishnan Mani[1][0009-0005-0731-8534]

[1] Department of Applied AI, Abhira, India.
[2] Faculty of Engineering, University of Sydney, NSW, Australia.

contactabhisheksebastian@gmail.com

**Abstract.** Fiber Specklegram Sensors (FSS) are vital for environmental monitoring due to their high temperature sensitivity, but their complex data poses challenges for predictive models. This study introduces Prion-ViT, a prion-inspired Vision Transformer model, inspired by biological prion memory mechanisms, to improve long-term dependency modeling and temperature prediction accuracy using FSS data. Prion-ViT leverages a persistent memory state to retain and propagate key features across layers, reducing mean absolute error (MAE) to 0.71°C and outperforming models like ResNet, Inception Net V2, and Standard Vision Transformers. This paper also discusses Explainable AI (XAI) techniques, providing a perspective on specklegrams through attention and saliency maps, which highlight key regions contributing to predictions.

**Keywords:** Fiber Specklegram Sensors (FSS), Prion Memory Mechanism, Vision Transformers (ViT), Explainability AI (XAI) and Temperature Prediction.

## 1 Introduction

Fiber Specklegram Sensors (FSS) are crucial for environmental monitoring due to their sensitivity to temperature and other variables, detecting minute changes through interference patterns in multimode optical fibers [1]. However, the complex, nonlinear nature of specklegram data poses challenges for predictive modeling, as traditional machine learning approaches, like CNNs, struggle to capture the global phase relationships and intensity variations in FSS patterns [2]. Despite advancements in deep learning, transformer-based architectures, such as Vision Transformers (ViTs) [3] and Swin Transformers [4], remain underutilized for FSS temperature prediction, with issues like gradient vanishing, information dilution, and difficulty capturing specklegram details affecting predictive performance [5][6].

This study introduces a novel approach to addressing these limitations by drawing inspiration from prion biology. In prion biology [7][8], certain proteins can exist in stable, misfolded states **(Fig. 1(a))** that can propagate similar conformational changes in other proteins **(Fig. 1(b))**, effectively encoding long-term memory [9][10]. This transformation does not occur simultaneously; rather, it is triggered by specific biological signals **(Fig. 1(c))**. In other words, when there is a need to store information, a signal acts as a gating mechanism that induces normal proteins to misfold. This phenomenon is particularly well-suited for modeling long-term dependencies, as prion-inspired memory mechanisms enable the retention and propagation of essential information for longer periods (in a machine learning analogy, across deeper layers). This approach can effectively overcome the limitations of traditional deep learning techniques in temperature prediction tasks.

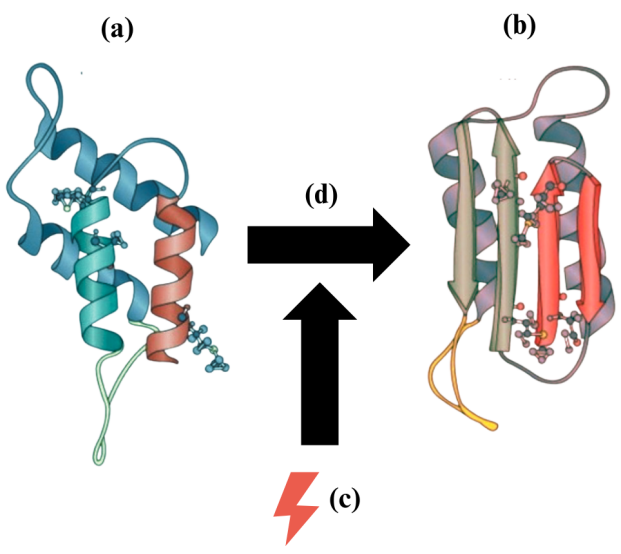

**Fig. 1.** (a) Illustrates the normal cellular form of the prion protein. (b) Represents the misfolded, infectious form of the prion protein. (c) Symbolizes the trigger or stimulus that causes the prion to misfold. (d) Indicates the conversion process from the normal prion protein to the misfolded form effectively encoding long-term information.

This paper introduces a novel model architecture called Prion-ViT, which integrates prion-like memory mechanisms into a Vision Transformer for forecasting temperature. The key innovation lies in the introduction of a persistent memory state within each transformer layer, enabling the model to retain crucial features across training epochs. This approach results in improved performance, with a Mean Absolute Error (MAE) of 0.71°C, surpassing the performance of conventional architectures such as ResNet and Inception Net V2.

The remainder of this paper is structured as follows: Section 2 reviews recent advancements in FSS technology and ViT-based models, Section 3 details the materials and methods used in this study, including data preprocessing and experimental setup, and Section 4 describes the Prion-ViT architecture and its theoretical justification. In Section 5, we present experimental results, including a discussion on Explainable AI (XAI) techniques such as attention maps and saliency maps, which provide insights into the model's focus and decision-making process for specklegram data. Section 6 concludes the paper, highlighting the model's potential for further research in optical sensing applications.

## 2 Literature Survey

Optical sensing technology, particularly Fiber Specklegram Sensors (FSS), has demonstrated diverse applications, including refractive index (RI) detection, temperature monitoring, water leak localization, torsion measurement, and pressure sensing. For example, Al Zain et al. [11] developed a highly sensitive multimode fiber (MMF) specklegram RI sensor using a 33 μm microfiber, achieving a temperature sensitivity of $0.013°C^{-1}$ between 25°C and 65°C via zero-mean normalized cross-correlation (ZNCC) analysis. Wang et al. [15] further demonstrated the utility of FSS in temperature sensing by correlating specklegram patterns with temperature changes in an MMF-single mode fiber (SMF) splice system. Inalegwu et al. [12] achieved 100% accuracy in detecting water leaks through convolutional neural networks (CNNs) for specklegram analysis, although minor accuracy declines were observed under varying conditions. Additionally, Li et al. [14] utilized ResNet to create a torsion sensor, achieving a prediction error of ±2° across a 0 – 360° range. Vélez-Hoyos et al. [13] optimized FSS for temperature sensing, achieving performance comparable to that of Fiber Bragg Grating (FBG) sensors. Other studies have expanded the versatility of FSS, such as Wang et al. [16] enabling simultaneous measurement of temperature and weight, Musin et al. [17] developing a modal interference-based surface temperature sensor, and Brestovacki et al. [18] using Raspberry Pi technology for cost-effective mechanical deformation sensing. Reja et al. [19] employed a multilayer perceptron (MLP) model for pressure sensing, achieving error rates below 3 kPa over a 1 MPa range. Collectively, these studies underscore the adaptability of FSS, especially when integrated with advanced machine learning techniques.

Simultaneously, Vision Transformers (ViTs) have emerged as strong alternatives to traditional CNNs in computer vision tasks. Unlike CNNs, which rely on convolutional layers for localized feature extraction, ViTs use self-attention mechanisms to capture global dependencies within images. Ovadia et al. [20] introduced the Vision Transformer-Operator (ViTO), a hybrid model that combines ViTs with operator learning to solve inverse partial differential equations (PDEs), demonstrating superior performance in super-resolution tasks compared to conventional operator networks.

Addressing challenges in dense prediction tasks, Xia et al. [21] developed ViT-CoMer, which improves spatial information exchange through multi-scale feature interactions. Dehghani et al. [22] introduced NaViT, a model that accommodates varying aspect ratios and resolutions without requiring image resizing. In addition, Wu et al. [23] proposed MeMViT, a model designed to enable efficient long-term video modeling with minimal computational cost, while Liu et al. [24] presented EfficientViT, which reduces computational load and enhances attention diversity through a cascaded group attention module. Abhishek et al. [27] explored a different approach to reduce computational costs in Vision Transformers by adding linear projection layers for plant leaf detection. Their research demonstrated a significant reduction in computation for edge computing applications in plant disease detection.

Despite these advancements in FSS and machine learning, a significant research gap remains in applying transformer-based architectures to temperature prediction tasks using specklegrams. This study seeks to bridge this gap by integrating Vision Transformers (ViTs), both with and without prion-based memory mechanisms, to improve long-term dependency handling. By examining how these models can effectively channel memory and address the limitations faced by conventional computer vision models [28], we aim to enhance both accuracy and adaptability in optical sensing applications.

## 3 Materials and Methods

### 3.1 Description of the dataset used in the study

The synthetic dataset [26] comprises 601 speckle patterns generated by modal interference in a multimode optical fiber, capturing temperature-induced variations from 0 °C to 120 °C in 0.2 °C increments. Modal interference occurs when multiple modes with distinct phase velocities and propagation constants overlap, causing constructive and destructive interference across the fiber's cross-section and producing observable speckle patterns. The experimental setup **(Fig. 2a.)** uses a 1490 nm semiconductor laser coupled into the multimode fiber via a tail fiber and flange. A heater precisely controls the fiber's temperature, altering its refractive index and dimensions. An industrial CCD camera captures the speckle patterns, which are then computationally analyzed to study temperature-induced modal interference effects.

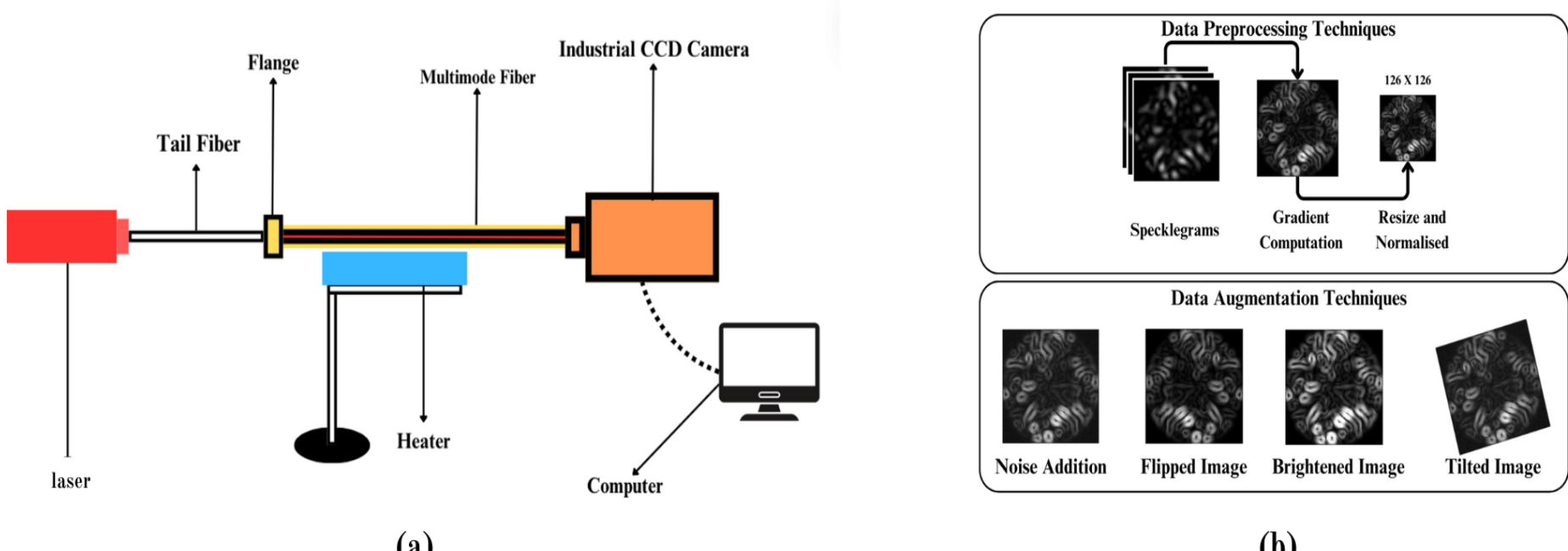

(a) (b)

Fig. 2. (a) Schematic of the experimental setup for dataset collection; (b) Preprocessing steps involved in the study.

### 3.2 Challenges in Analyzing Specklegram Images with Deep Learning

Analyzing specklegram images is challenging due to their complex, random optical interference patterns, where temperature variations cause subtle shifts in intensity distributions, and fine-scale noise obscures meaningful patterns for CNNs. To address these issues, we applied preprocessing techniques such as Sobel gradient computation, normalization, and resizing, which highlight rapid intensity changes and enhance underlying patterns for better deep learning performance. Additionally, data augmentation methods **(Fig. 2b.)** like noise addition, image flipping, brightness adjustment, and rotation increased input variability, improved model robustness, and enhanced generalization. This comprehensive strategy refined feature extraction, enabling accurate temperature prediction from complex specklegram patterns.

# 4 PROPOSED METHODOLOGY

In our methodology, we utilized a preprocessed dataset to train both pre-trained models and custom architectures. We validated the accuracy and reliability of our results using quantitative metrics such as MSE, MAE, RMSE, Maximum Error, and $R^2$ score. Each model was consistently trained for 100 epochs with a data split of 70% for training, 20% for testing, and 10% for validation. The training process was conducted on an A100 GPU using Google Cloud Colaboratory Notebooks, ensuring efficient computation and standardized performance evaluation. This setup enabled robust comparisons across models and facilitated a thorough assessment of their predictive capabilities.

## 4.1 Transfer Learning with Pre-trained models (Mobile Net, Inception Net V2 & ResNet).

MobileNet, chosen for its computational efficiency, struggles with Fiber Specklegram Sensor (FSS) data due to its design for large-scale image classification, which emphasizes high-level object features over the fine, continuous variations present in specklegrams. Its depth-wise separable convolutions are inadequate for capturing the high-dimensional, nuanced features of FSS, where intricate spatial and modal interactions respond sensitively to minor environmental changes. Similarly, Inception net V2, although adept at multi-scale feature extraction, exhibited higher MAE and MSE in temperature predictions for FSS data **(See (Fig 7a.))**, as its parallel convolutional layers could not adequately model the fine-grained interference patterns. In contrast, ResNet, with its residual connections, performed better by enabling deeper network training and extracting intricate features, outperforming both MobileNet and Inception net V2 **(See (Fig 7a.))**. However, ResNet still fell short of achieving the high precision necessary for accurate temperature predictions in specklegram-based sensing, highlighting the need for more specialized architectures that incorporate mechanisms like attention or spectral-domain processing to effectively interpret the unique optical and modal characteristics of specklegram data.

## 4.2 Vision Transformers and Swin Transformers.

The Vision Transformer (ViT) model, adapted from NLP transformers for image tasks, was designed to address the challenges of FSS data. Instead of using a transfer learning approach with Google's pre-trained ViT model (ViT-base-patch16-224) [3], we also utilized Swin Transformers — an advanced variant of Vision Transformers (ViTs) optimized for FSS applications. Unlike the original Swin Transformer models [4], which were designed by Liu et al. for classification tasks and trained on the ImageNet dataset, our adaptations of both models **(Fig. 3a. & 3b.)** were built from the ground up. This was essential because training of the base models on ImageNet 1k dataset does not align with the specific requirements of specklegrams.

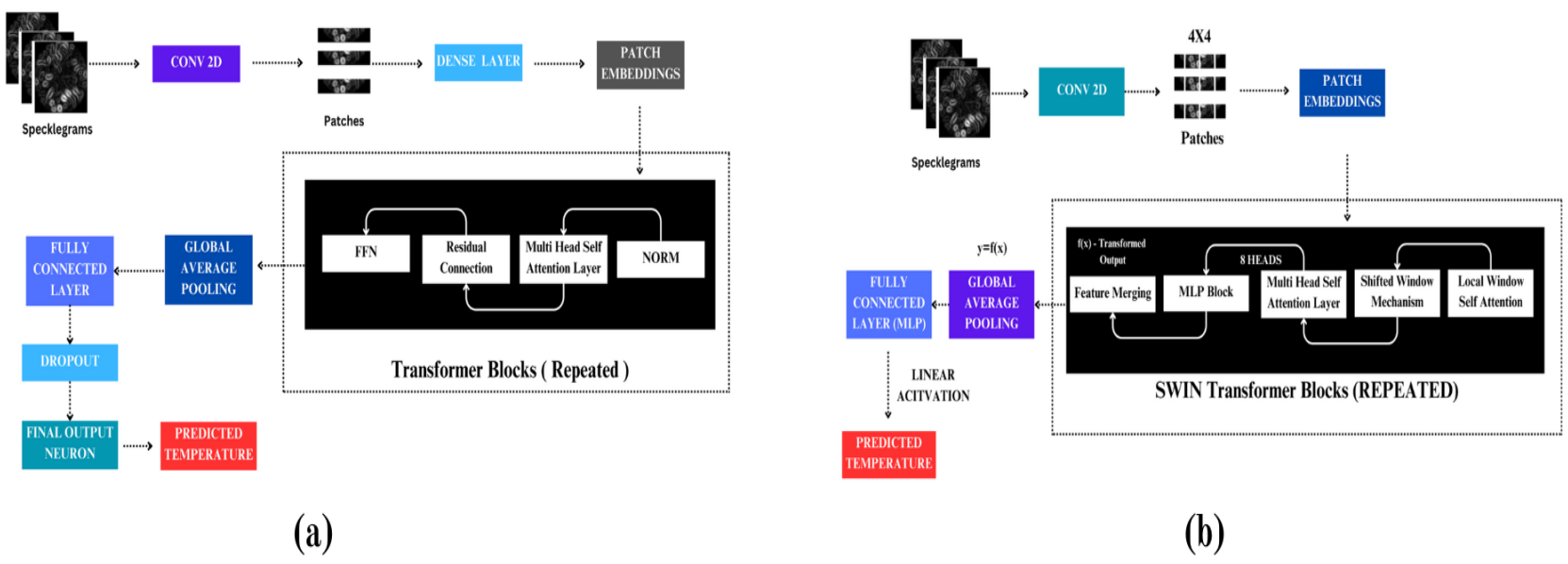

**Fig. 3.** Architecture of vision transformers and swin transformers for temperature prediction.

The ViT model processes input specklegrams of size (126, 126, 3) by dividing them into 16×16-pixel patches via a Conv2D layer, transforming the 2D image into a sequence analogous to words in NLP. These patches are flattened and mapped to 64-dimensional embeddings, with positional embeddings added to retain spatial information crucial for interpreting interference fringes. The core comprises four transformer blocks, each featuring layer normalization and multihead self-attention to capture dependencies between patches, residual connections to prevent vanishing gradients, and a feed-forward network with two dense layers (expanding to 128 dimensions and then reducing back). Post-transformer layers include global average pooling, a fully connected layer with 2,048 neurons and ReLU activation, dropout (0.5) to prevent overfitting, and a final linear layer for temperature prediction. However, the primary difference in architecture between Swin Transformers and ViTs is that, instead of using global attention like in ViTs, Swin Transformers apply self-attention within 4×4-pixel local windows. This reduces computational complexity and enhances spatial dependency capture through a shifted window technique.

## 4.3 Introducing Different Blocks Memory to Vision Transformers.

### 4.3.1 Theoretical Rationale Over the need of a better Memory Unit Than LSTM and GRU.

LSTM networks, with their complex architecture and high computational demands, are slow and inefficient, especially for large datasets or deep architectures. GRUs, simpler than LSTMs, also face similar challenges. Both LSTMs and GRUs overfit on small datasets and suffer from the vanishing gradients problem, limiting their ability to retain long-term dependencies. Their sequential processing restricts parallelization, leading to inefficient training. To address these limitations, a memory unit should reduce computational complexity, minimize overfitting, handle long-term dependencies effectively, and support parallel processing. Prion memory blocks, inspired by prion-like proteins, offer the robustness and efficiency that LSTMs and GRUs lack for speckle-gram-based temperature prediction.

Prion-ViT enhances Vision Transformers (ViTs) by embedding a persistent memory state with each transformer layer **(Fig. 4a.)**. This biologically motivated system enables the model to retain and propagate crucial information across layers efficiently, avoiding the vanishing gradient issue. Prion-ViT's core innovation is the integration of prion memory blocks, which emulate the stability-maintaining behavior of prion proteins. This allows ViTs to retain, update, and propagate essential information consistently. Theoretical justification includes the ability to maintain a consistent memory state, preserving important features across deep layers. Batch dimension averaging provides regularization, capturing common patterns and minimizing random noise. Additionally, the gating mechanism adaptively controls information flow, like the forget gates in LSTMs, further improving long-term dependency modeling. This makes Prion-ViT effective for complex tasks like temperature prediction in high-dimensional Fiber Specklegram Sensor data.

### 4.3.2 Mathematical Formulation and Integration of Prion Memory blocks.

Let:

- $X^{(l)} \in \mathbb{R}^{B\ x\ N\ x\ D}$ be the output of the $l^{th}$ transformer block, where $B$ is the batch size, $N$ is the number of patches (tokens), and $D$ is the embedding dimension (*such as specific structures or states in prions*).
- $M \in \mathbb{R}^{N\ x\ D}$ be the shared memory state that persists across batches and transformer blocks (*such as stable conformations in prions*).
- $G^{(l)} \in \mathbb{R}^{B\ x\ N\ x\ D}$ be the gate values computed at the $l^{th}$ block (*such as activation commands in prions*).

***1) Gate Computation.***

The gate values $G^{(l)}$ are computed by applying a learned transformation to the output $X^{(l)}$ as in equation 1.

$$G^{(l)} = \sigma\left(X^{(l)} W_g + b_g\right) \qquad (1)$$

Where,

- $W_g \in \mathbb{R}^{D\ x\ D}$ and $b_g \in \mathbb{R}^{D\ x\ D}$ are learnable parameters.
- $\sigma$ is the sigmoid activation function, which outputs values in the range (0,1), indicating how much of the memory should be retained.

***2) Memory Update Rule.***

The memory $M$ is updated based on the gated combination of the previous memory state and the new output from the transformer block as in equation 2.

$$M = \frac{1}{B}\sum_{b=1}^{B}(G_b^{(l)} \odot \, M \, + (1 - G_b^{(l)}) \odot X_b^{(l)}) \qquad (2)$$

Where,

- $\odot$ denotes element-wise multiplication.
- $G_b^{(l)}$ is the gate value for the $b^{\text{th}}$ sample in the batch.
- $X_b^{(l)}$ is the transformer block output for the $b^{\text{th}}$ sample in the batch.

This update rule allows the memory to retain relevant features while incorporating new information, controlled by the gate values.

### *3) Memory Broadcast.*

Once the memory $M$ is updated, it is broadcast across the batch for use in the next transformer block as in equation 3.

$$\widetilde{M^{(l)}} = 1_{BX1X1} \otimes M \qquad (3)$$

Where,

- $1_{BX1X1}$ is a tensor of ones used to replicate M across the batch dimension.

### *4) Input to Next Transformer Block.*

The updated memory $M$ is then used as the input to the next transformer block as in equation 4.

$$X^{(l+1)} = \widetilde{M^{(l)}} \qquad (4)$$

This process allows the memory to persist across layers, continually updating based on the output of each transformer block.

The Prion Memory Mechanism can be integrated into the standard ViT architecture by adding the memory update and gate computation steps between transformer layers as in Algorithm 1.

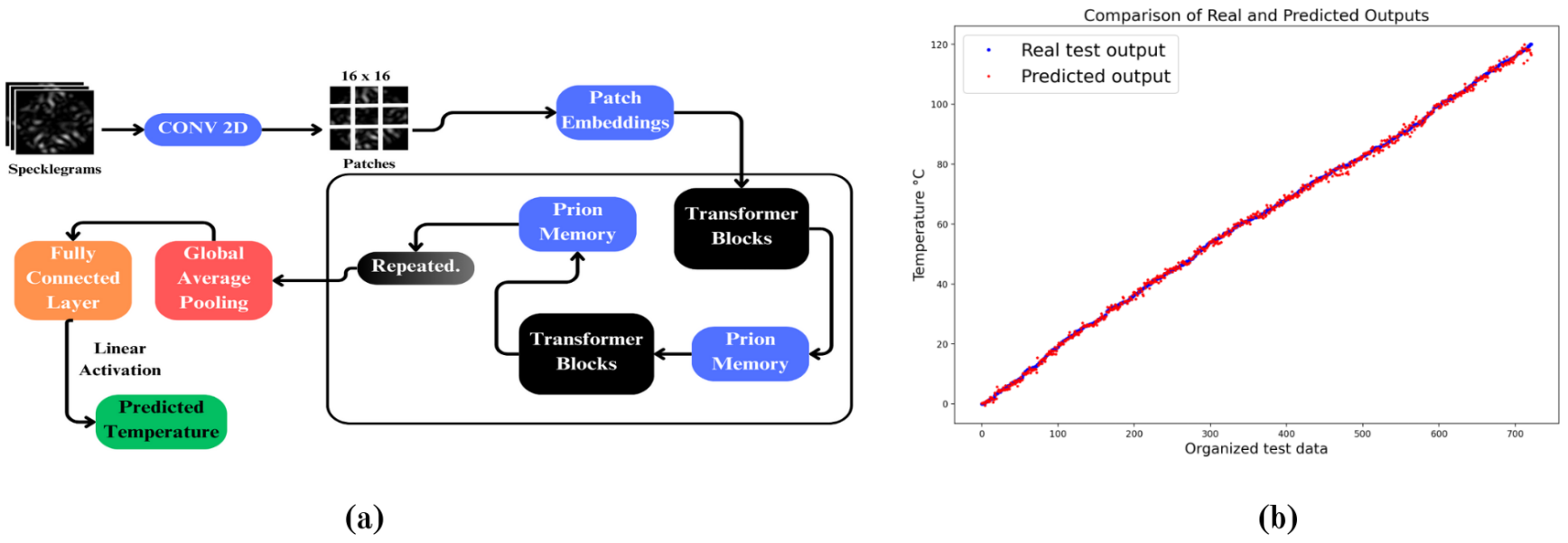

**Fig. 4.** (a)Architecture of Prion-ViT for regression, (b) Test Results of Prion-ViT model.

**Algorithm 1: Prion Memory Mechanism**

1. **Initialization**: Initialize the memory state $M$ to zeros: $M = 0_{NxD}$
2. **For each transformer block $l$=1 to L:**
   a. Compute gate values: $G^{(l)} = \sigma(X^{(l)}W_g + b_g)$
   b. Update memory for each batch sample $b$: $M_b^{(l)} = G_b^{(l)} \odot M + (1\text{-}G_b^{(l)}) \odot X_b^{(l)}$
   c. Aggregate the updated memories across the batch: $M = \frac{1}{B}\sum_{b=1}^{B} M_b^{(l)}$
   d. Broadcast the updated memory: $X^{(l+1)} = 1_{BX1X1} \otimes M$

# 5 RESULTS AND DISCUSSIONS

The Prion-ViT model's performance was evaluated against various architectures[2][25], **(See (Fig 5a.))** including ViTs, Swin Transformers, CNN+ANN , ANN, MobileNet, Inception Net V2, and ResNet, using FSS test data. The comparison metrics included Mean Squared Error (MSE), Mean Absolute Error (MAE), Root Mean Squared Error (RMSE), Maximum Error, and R² Score.

Prion-ViT outperformed the remaining models in processing FSS data, capturing complex optical interference patterns in specklegrams more effectively **(See (Fig 4b.))**. Unlike traditional models, Prion-ViT's unique Prion memory blocks enable it to retain intricate modal interactions across layers, simulating stable information storage found in biological systems. This adaptation allows Prion-ViT to better model the nuanced and continuous changes in speckle patterns associated with temperature fluctuations.

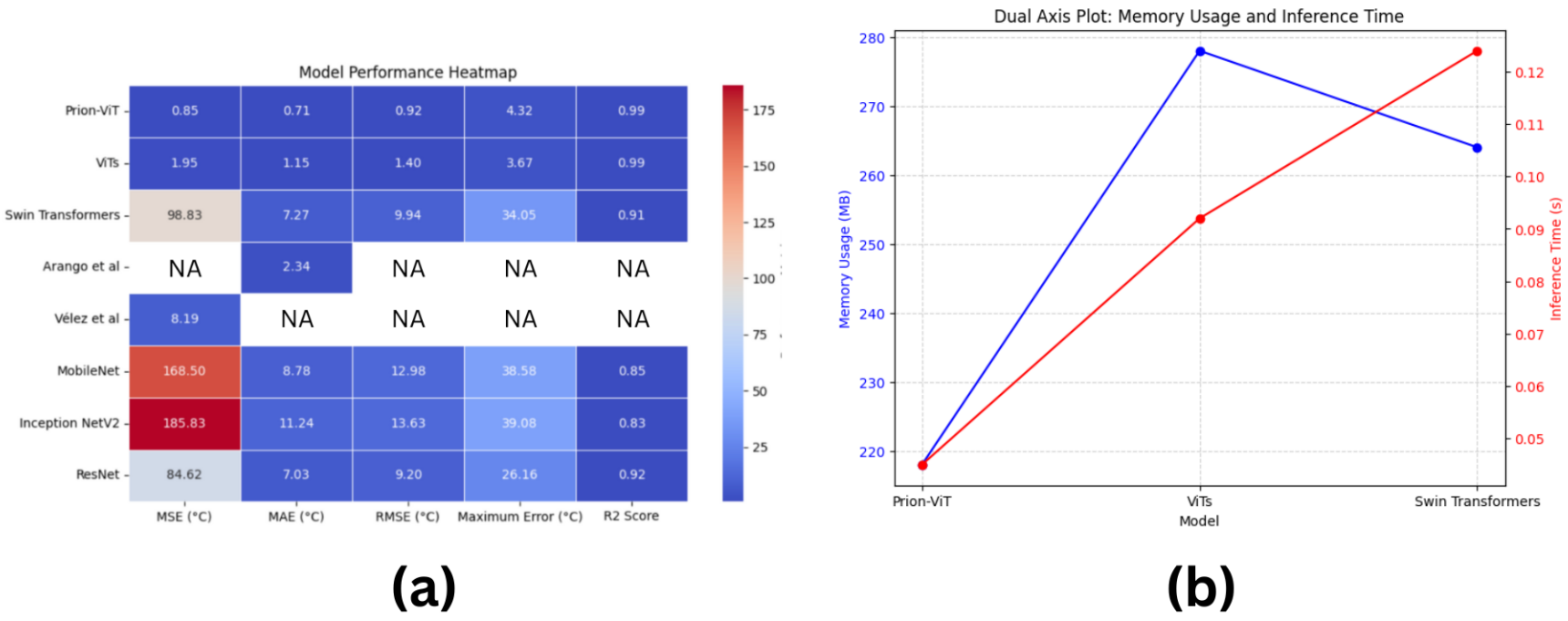

**Fig. 5.** (a) Comparative Analysis of Different Models, (b) Dual Axis Plot of Memory Usage and Inference Time of Different Vision Transformer Models.

An ablation study comparing ViT with and without Prion Memory blocks confirmed that these blocks significantly enhance model performance, reducing the Mean Absolute Error from 1.15°C to 0.71°C. Swin Transformers, which rely on shifted windows for local dependency handling, struggled with the granular interference patterns of specklegrams. CNN+ANN and ANN models were also less effective, as they require defined edges and repetitive textures, which specklegram images lack.

Optimized for efficiency in large-scale classification, MobileNet and Inception Net V2 were unsuitable for handling the fine-scale variations in specklegram data, resulting in high prediction errors. ResNet performed moderately better due to residual connections, yet was unable to match Prion-ViT's specialized abilities for FSS data.

Prion-ViT demonstrates superior memory and inference efficiency **(Fig 5b.)**, attributed to its Prion memory blocks that maintain an optimized memory state, reducing redundant computations. Faster inference times also indicate Prion-ViT's suitability for real-time applications requiring computational efficiency.

The model's ability to leverage past contextual information allows Prion-ViT to adapt to dynamic data distributions, capturing the intricate temporal dependencies necessary for complex pattern recognition tasks such as specklegram-based temperature sensing. This adaptability makes it an excellent candidate for real-time optical sensing, where precise temporal modeling is critical.

Prion-ViT also enhances explainability (XAI) **(Fig 6.)** through attention maps and saliency maps, which provide a visual understanding of the model's focus during inference. The attention maps **(Fig 6a.)** highlight how the model distributes focus across different regions of the input specklegrams, while saliency maps **(Fig 6b.)** pinpoint the most influential areas contributing to the predictions. This transparency aids in interpreting the model's decisions, validating its robustness, and fostering trust in real-time optical sensing applications where precise temporal and spatial modeling is critical.

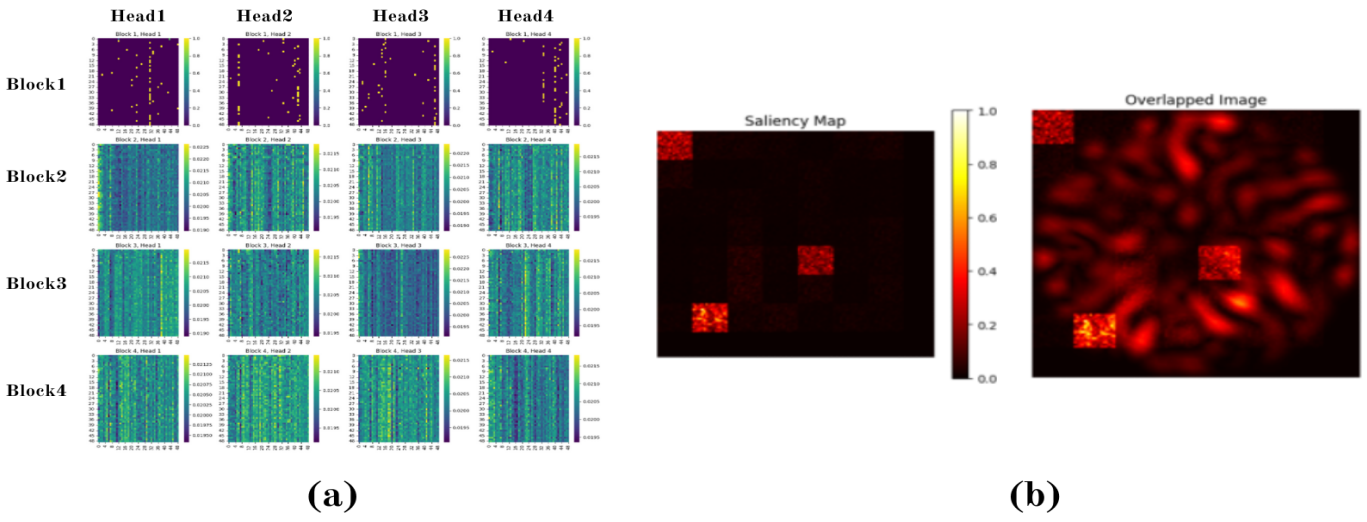

**Fig. 6.** (a) **Attention Maps**: Visual representation of the attention weights in a Vision Transformer, showing the distribution across multiple heads (Head1–Head4) and blocks (Block1–Block4). Darker or lighter regions signify the degree of focus across the input features. (b) **Saliency Maps**: Heatmaps highlighting regions contributing most to the model's predictions. The overlapped image combines the saliency map with the original specklegram, emphasizing key areas detected by the model. These visualizations help interpret the model's behavior, showing where the model focuses during decision-making, which is critical for understanding its performance on specklegram-based temperature prediction tasks.

In **Figure 7**, we illustrate a practical application of an underwater fiber temperature monitoring system leveraging fiber specklegrams. This system involves deploying optical fibers along the ocean floor, with poles equipped with communication modules and cameras to facilitate monitoring and data transmission. By utilizing the fiber specklegram technique, the setup detects variations in light patterns within the optical fibers, allowing for accurate temperature measurements in underwater settings.

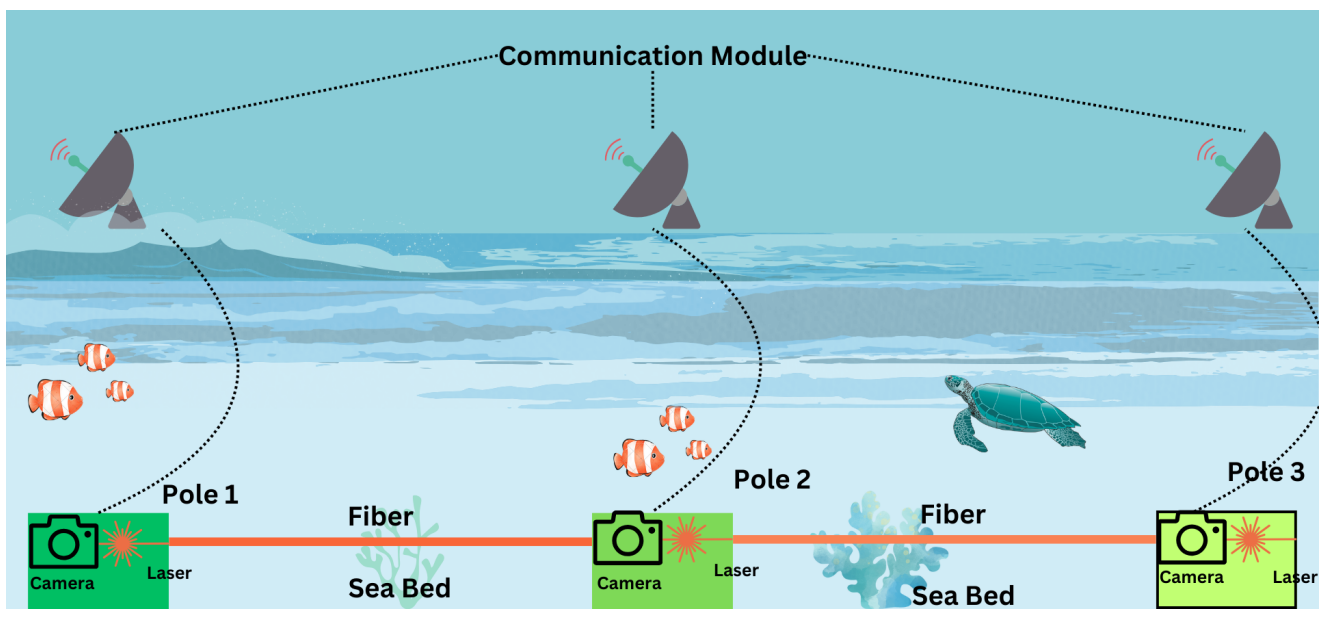

**Fig. 7.** Envisioning a real-world implementation of an underwater fiber temperature reading system using fiber specklegram sensors.

This integration of cutting-edge transformer architectures and innovative optical sensing techniques underscores the potential for substantial advancements in temperature monitoring and related applications **(See Figure 7.)**.

## 6 CONCLUSION

This study presents Prion-ViT, a prion-inspired Vision Transformer for temperature prediction using Fiber Specklegram Sensor (FSS) data. The novel Prion Memory Mechanism retains critical features across layers, addressing limitations of LSTM and GRU. Prion-ViT reduced Mean Absolute Error to 0.71°C, outperforming ResNet, Inception Net V2,Vision Transformers and Swin Transformers. Key contributions include enhanced long-term memory retention, improved adaptability to small datasets, and selective transformer block dropout for better generalization. Future research could optimize Prion-ViT's efficiency, apply it to other optical sensing tasks (such as pressure sensing), use advanced data augmentation, and integrate it with CNNs or GNNs to enhance feature extraction in complex, high-dimensional data.